\documentclass[graybox]{svmult}
\usepackage{times}
\usepackage{latexsym}
\usepackage{graphicx}
\usepackage{algorithm}
\usepackage{comment}
\usepackage[capitalise, noabbrev]{cleveref}
\usepackage{comment}
\usepackage{multirow}
\usepackage{svg}

\usepackage{algorithm}
\usepackage[noend]{algpseudocode}

\usepackage{microtype}

\algnewcommand\algorithmicforeach{\textbf{for each}}
\algdef{S}[FOR]{ForEach}[1]{\algorithmicforeach\ #1\ \algorithmicdo}

\title*{Towards Handling Unconstrained User Preferences in Dialogue}

\author{Suraj Pandey, Svetlana Stoyanchev and Rama Doddipatla}
  
 \institute{Suraj Pandey \at The Open University, Walton Hall, Kents Hill, Milton Keynes MK7 6AA, \email{suraj.pandey@open.ac.uk}
\and Svetlana Stoyanchev, Rama Doddipatla \at Toshiba Europe Ltd., Cambridge Research Laboratory, 208 Cambridge Science Park Milton Rd, Cambridge CB4 0GZ \email{svetlana.stoyanchev,rama.doddipatla@crl.toshiba.co.uk}}

\date{}

\begin{document}
\maketitle
\abstract{
A user input to a schema-driven dialogue information navigation system, such as venue search, is typically constrained by the underlying database which restricts the user to specify a predefined set of preferences, or slots, corresponding to the database fields. We envision a more natural information navigation dialogue interface  where a user has flexibility to specify unconstrained preferences that may not match a predefined schema. We propose to use information retrieval from unstructured knowledge to identify entities relevant  to a user request.  We update the Cambridge restaurants database with unstructured knowledge snippets (reviews and information from the web) for each of the restaurants and annotate a set of query-snippet pairs with a relevance label. We use the annotated dataset to train and evaluate snippet relevance classifiers, as a proxy to evaluating recommendation accuracy.
We show that with  a pretrained transformer model as an encoder, an unsupervised/supervised classifier achieves a weighted F1 of .661/.856.}

\section{Introduction}

\begin{figure}[t!]
    \centering
     \caption{Process for building a supervised relevance scoring model. }
    \includegraphics[width=0.8\textwidth,scale=0.5]{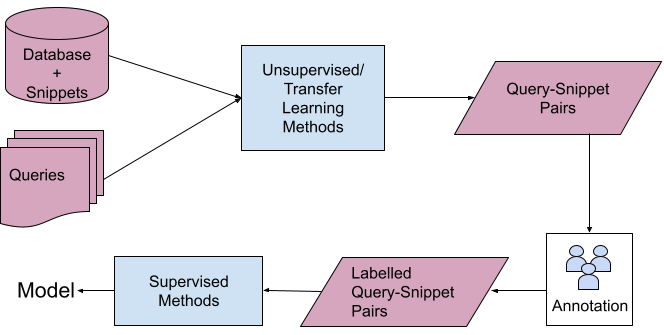}
   
    \label{fig:method:process}
\end{figure}

A conversation is a natural user interface for accessing information.
In {\it information navigation} dialogue, such as search (for restaurants, hotels, or tourist attractions), a user specifies search constraints and navigates over search results using text-based, spoken, or multi-modal interface. Information navigation tasks are typically handled with a schema-driven approach~\cite{budzianowski-etal-2018-multiwoz,schema-guided,stoyanchev-asu-2021}. While a schema-driven system may be effectively bootstrapped for a new application using the structure and content of the corresponding database, a user is limited in the range of constraints they may specify.
For example, in a restaurant search domain with the schema fields {\em area}, {\em  price range} and {\em food type}, a user may specify any combination of these fields but not others~\cite{henderson-thomson-williams:2014:W14-43}. 
To evaluate a schema-driven information navigation system, the recruited experiment participants are typically given a conversation `goal' based on the database schema, e.g. {\em ``you are looking for a cheap Italian place in the center"}, and are instructed to retrieve matching venues using spoken or text chat interaction. These instructions guide the user to specify in-domain preferences that can be handled by the dialogue system. 
An initial user request based on the above `goal'   may be: 
\begin{quote}
User: {\em `A cheap Italian restaurant in the centre'} \\
User: {\em `I am looking for a cheap place'}\\
User: {\em `A restaurant in the center'}\\
User: {\em `An Italian restaurant in cheap price range'}\footnote{A system may ask to narrow down the search criteria and a user may specify additional preferences in consecutive turns.}\\
User: {\em ...}
\end{quote} 
While there is variability in the natural language utterances of a recruited user for a predefined `goal', the preference type specified in the goal is limited to the schema. 

Unlike the recruited subjects, real users come up with personalised search preferences and formulate their requests to the system without a bias of the instructions. The challenge is that a user request with preferences outside of the domain schema, e.g. {\em  `Find me a cosy family friendly pub that serves pizza.'}  can not be handled by a purely schema-driven system. The majority of user queries (75 out of 105) that we collected without priming the user by a predefined `goal' do not mention any of the domain-specific schema fields, indicating that a purely schema-driven system is not sufficient to handle naturally constructed user requests. 
In contrast to a schema-driven dialogue system, a search interface handles such unconstrained user queries using information retrieval methods from unstructured (text) data. Search interfaces, however, are not interactive and do not handle query changes or follow-up questions. In this work,  we aim to {\bf improve the naturalness of interaction } with an information navigation dialogue system. We propose to {\bf extend the  schema-driven dialogue system to handle out-of-schema user queries} by incorporating an entity retrieval module in the dialogue  system pipeline.

An entity retrieval model requires domain-specific data to extract the requested information.  For this study, we create an annotated dataset for the Cambridge Restaurants domain following the process outlined in~\cref{fig:method:process}.
We first collect a database from the Web, including text snippets composed from restaurant reviews and descriptions along with a set of unconstrained restaurant search queries. Next,  unsupervised and transfer learning methods are used to obtain a set of query-snippet pairs which are then annotated using Amazon Mechanical Turk. The resulting annotated dataset is then used to build supervised relevance scoring models and compared with the performance of unsupervised and transfer learning approaches. We show that using a pretrained transformer model as an encoder, an {\bf unsupervised/supervised classifier of the snippet relevance to the query achieves a weighted F1 of .661/.856}. 

The contributions of this paper are:
\begin{itemize}
    \item A methodology for extending a schema-driven dialogue system to support natural user preferences. 
    \item A manually annotated dataset with 1.7K query and text snippet pairs. 
    \item Evaluation of supervised, unsupervised, and transfer learning approaches for snippet relevance classification.
    
\end{itemize}

The rest of the paper is organized as follows. In~\cref{sec:relwork}, we outline related research. In~\cref{sec:data}, we present an extended Cambridge Restaurants 2021 dataset and annotated REStaurant Query Snippet dataset (ResQS). In ~\cref{sec:method},  we describe the approaches used to detect text snippets relevant to a query and present experimental results in~\cref{sec:exp}. Finally, the conclusions are presented  in~\cref{sec:conclusions}.

\section{Related Work}
\label{sec:relwork}
 
Conversational search aims at providing users with an interactive natural language  interface for access to unstructured open-domain information~\cite{trippas2018informing}. Similarly, task-oriented information navigation dialogue systems often involve search, e.g. for venues or catalogue items ~\cite{budzianowski-etal-2018-multiwoz,10.5555/3298023.3298238}. 
While open-domain conversational search has a wider scope than  closed-domain search in task-oriented dialogue, empirical analysis shows that both tasks result in a similar conversational structure
~\cite{10.1007/978-3-030-15712-8_35}.
 
Open-domain search interfaces and task-oriented information navigation dialogue systems both take unconstrained natural language as input. However, dialogue system users are typically limited in expressing their preferences by the domain schema. In response to an out-of-schema user request, a task-oriented dialogue system may produce an informative help message guiding the user to adapt to its limitations 
~\cite{DBLP:journals/umuai/KomataniUKO05,tomko-rosenfeld-2004-speech}. Alternatively, system capabilities may be extended beyond a domain API. For example, Kim et al. \cite{kim-etal-2020-beyond} proposes a method for handling users' follow-up questions in task-oriented dialogue systems. To support pragmatic interpretation, Louis at al. \cite{louis-etal-2020-id} explores users' indirect responses to questions. To extend a task-oriented system to handle natural preferences, a corpus of natural requests for movie preferences was collected using a novel approach to preference elicitation \cite{Radlinski2019}.  

Task-oriented dialogue systems require accurate models to extract information from unstructured text. Pretrained transformer models, such as BERT~\cite{devlin2019bert}, have shown to be effective in extracting information from text, leading to significant improvements on
many NLP tasks, including open-domain question answering, FAQ retrieval, and dialogue generation \cite{wang2019multi,sakata2019faq,kim-etal-2020-beyond}. 
 Following previous work, we use BERT both in a supervised and an unsupervised setting~\cite{izacard2021leveraging, 10.1145/3397271.3401325}. 
We also explore transfer learning from a general natural language entailment task using publicly available corpora~\cite{conf/emnlp/BowmanAPM15}.

\section{Data}
\label{sec:data}
To develop and evaluate an entity retrieval component for a dialogue system that handles unconstrained user queries, it is necessary to construct a dataset that includes text snippets associated with the entities, collect unconstrained user queries, and identify a set of matching entities for the queries. We collect a new extended dataset of Cambridge restaurants (\cref{sec:data:camdb}), a set of unconstrained search queries (\cref{sec:data:queries}), and annotate query-snippet pairs with relevance labels~(\cref{sec:data:annotation}).

\subsection{Cambridge Restaurants 2021}
\label{sec:data:camdb}

In previous work~\cite{henderson-thomson-williams:2014:W14-43}, the authors used a now outdated dataset of 102 Cambridge restaurants, which did not have review information.
In this work, we created an up-to-date database with 422 restaurants in Cambridge, UK.\footnote{The dataset is compiled by crawling the Web in January 2021.} Following the schema  used in previous work, each restaurant is associated with {\em cuisine, price range, location,} and {\em description}. As in the past systems, the price range is mapped to {\em cheap, moderate, expensive} and location to {\em east, west, centre, south}. However, {\em cuisine} in our database is associated with a list of values rather than a single value for each entity. In addition, the new dataset includes information on {\em meals} (breakfast, lunch, dinner), {\em special diets}  (e.g., vegan, gluten free) and {\em reviews}.

A standard restaurant database may not always contain information relating to the unconstrained user's search preferences (example, {\it cosy family friendly pub}). In such regards, we theorise that personalised messages, such as  {\it reviews}, are an acceptable source to handle such requests. We collect 62.3K reviews  with an average and standard deviation of 145(256) per restaurant. The reviews together with text in each data fields are used as {\it text snippets} to retrieve items relevant to the query in the experiments. Only positive reviews (rating 4 or 5 stars) are used,  as we expect user queries to mention desirable properties of the restaurant. 

\subsection{Unconstrained Queries}
\label{sec:data:queries}

To simulate natural unbiased user requests in a restaurant search domain, we created an online form with one question:
{\it  Please type a sentence describing your restaurant preference to your smart virtual assistant.'}
The form was distributed to several university and company mailing lists.  Each participant was asked to enter one  query.\footnote {As the task was very short, the participants were not paid.}

We found that only 30 out of 105 collected queries specified a constraint corresponding to one of the pre-defined fields of the database schema (area, cuisine, or price range) that may be used to search for a restaurant in a purely schema-driven system. Although only 14 of these queries contained an entity exactly matching a value in the database and  only 7 had no other preferences besides the slot value. For example, {\it  `I would like to eat in a fine dining establishment, preferably \textbf{french} cuisine.'} contains a mention of cuisine as well as a vague preference for a {\it `fine dining establishment'}. The unbiased queries are highly diverse and most frequently contain a menu item (46), a subjective  (31) or an objective preference (25) about a restaurant (see~\cref{tab:data:qproperties}).
We removed 5 queries that mention proximity, e.g. {\it `near me'}. Such queries would not be handled with the proposed method as it would require additional geographical location information.  We use the remaining 100 queries in our experiments.

\begin{table*}[t!]
    \centering
    \caption{Preferences (number of queries) mentioned by the users in 105 unbiased queries. }
    \begin{tabular}{|p{2cm}|p{9cm}|}
    \hline

preference & examples \\ \hline\hline
menu item (46)& burger, veggie burger, pizza, curry, steak, fried rice, meat, sushi, seafood, noodle bar, dim-sum, shark, eel, alcohol, wine, mulled wine, beer, whisky,  milkshake, dessert\\\hline 
objective (25) & live music, music, quiet, dogs, kids, fireplace, free delivery, local food, new place, biggest pizza, spicy food, large groups, Burn's night, big pizza,sweet tooth, traditional, portion size, restaurant name, parking, quick, outdoor seating, spaced out tables, byob \\\hline
subjective (31) & ambience, friendly, gourmet, authentic, romantic, cozy, small, local, safe, different, inventive, stylish, exotic, interesting, hygienic, fine dining, best value, fancy, nothing too fancy\\\hline 

    \end{tabular}
    
    \label{tab:data:qproperties}
\end{table*}

\subsection{Query-Snippets Annotation}
\label{sec:data:annotation}
For extracting relevant entities using unstructured data, we need to develop supervised models, which require preferably a balanced training dataset.
We create a   REStaurant Query Snippet (ResQS) dataset of query-snippet pairs, where the snippets include text from reviews and each of the database fields, labelled with `1' if the snippet is {\it relevant } to the query and `0', otherwise. For each query, most of the snippets from our set of 62.3K candidates will be irrelevant. Thus, a random selection of snippets for each query would result in an unbalanced dataset of mostly irrelevant snippets.
Instead, unsupervised and transfer learning methods are used to select a set of relevant snippets for each query.  We then use manual annotation on the selected set to create query-snippet pairs with annotated relevance information. 

For each of the 100 queries, we first score all snippets using two unsupervised and one transfer learning method (see the first three methods shown in~\cref{fig:method:models}). 
Next, we  compute a relevance score for each of the 422 restaurants in the dataset and use it to rank  them (see~\cref{sec:method}). Finally, we randomly sample one of the  five top-ranked restaurants for the query\footnote{The number of top results (5) was chosen empirically since  a user of a dialogue system may navigate over multiple search results.} and manually label the top five snippets that were used to compute the relevance score of this restaurant. Manual labels are used for the evaluation of the unsupervised methods as well as for training the supervised models.  Using three methods to select five snippets per query we generate $1500$ query-snippet pairs.  
Additionally, to simulate the hybrid system where both relevance ranking and database match are used to extract a relevant item, we use the 14 queries that specify a value for one of the system slots (\emph{area}, \emph{cuisine}, or \emph{price range}). This subset is then used to select a recommended restaurant for the query. Thus by using three methods to select five snippets for the 14 queries we extend the dataset by $210$ query-snippet pairs resulting in $1710$ examples.

For annotation, we use crowd workers from Amazon Mechanical Turk. Each Human Intelligence Tasks (HIT) consists of 23 randomized query-snippet pairs and is annotated by three crowd workers.\footnote{3 query-snippet pairs with a known label are used to monitor work quality. The workers that did not pass the quality test were rejected.} The workers were compensated according to government pay guidance.\footnote{\url{https://www.gov.uk/national-minimum-wage-rates}} We use the majority vote amongst the three crowd workers to select the final label. The pairwise agreement between each annotator and the majority vote is very good with Fleiss Kappa~$k>.7$~\cite{fleiss1973equivalence}.

\begin{algorithm}[t!]
\caption{Relevance ranking}
\label{algo:method:rel-item}
\small
\begin{algorithmic}[1]
\Procedure{RankAndSelect}{$Query$,$~Dataset$, $J$, $N$}
    \State {\textbf{Input}: $Query$ is a String of user's request}
    \State {\textbf{Input}: $Dataset$ is a collection of items with associated snippets ${Item:[S1,S2,...]}$}
    \State {\textbf{Input}: $J$ is the number of top snippets used to compute the score}
    \State {\textbf{Input}: $N$ is the number of items to return}
    \State {\textbf{Output}: List of top $N$ items for the $Query$}

        \ForEach{$Item \in Dataset$}
            \ForEach{$Snippet \in Item$}
                \State $Snippets[score] = RelevanceScore( Query, Snippet)$
            \EndFor
            \State $Sorted\_Snippets = Sort(Snippets.score, order=descending)$ 
            
            \State${\displaystyle Item[score]={\frac {1}{J}}\sum _{i=1}^{J}Sorted\_Snippets\_i[score]}$
            
        \EndFor
        \State $Sorted\_Items = Sort(Items.score, order=descending)$ 
        \State \Return $[Sorted\_Items_1..Sorted\_Items_N]$
\EndProcedure

\end{algorithmic}
\end{algorithm}

\section{Method}
\label{sec:method}
In this work, we extend a schema-driven dialogue system to support users' unconstrained search requests. Our goal is to extract the items most relevant to the users' query. A dialogue system then presents these items as recommendations to the user and responds to follow-up questions. 
This work only addresses the extraction of the relevant items, leaving the evaluation of information presentation and handling of interaction to future work. We apply and evaluate our approach using the Cambridge Restaurants search domain.

 \cref{algo:method:rel-item} is used to extract an item most relevant to the user's query. First  the snippets are scored (line 9) then the top recommended restaurants are obtained by ranking the list of restaurants based on the average score of the top 5 snippets.\footnote{We use empirically chosen J=5 and N=5 in this work.}

\begin{figure}[t!]
    \centering
     \caption{Relevance scoring methods.}
    \includegraphics[width=\textwidth]{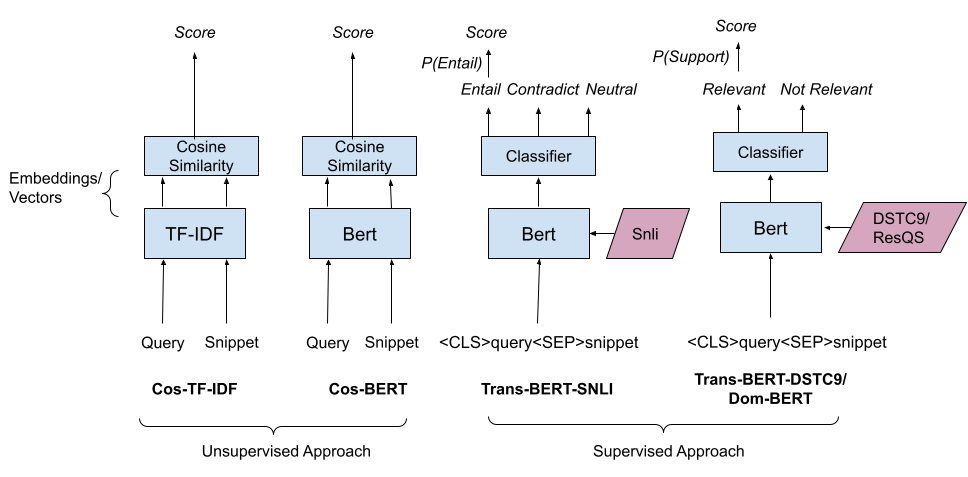}
   
    \label{fig:method:models}
\end{figure}

For snippets scoring, we explore two unsupervised approaches: Cos-TF-IDF and  Cos-BERT,  and two supervised approaches: transfer learning from another domain (Trans-BERT), and in-domain training approach (Dom-BERT) illustrated on~\cref{fig:method:models}. The input to each of the models are two strings (query and text snippet) and the output is a  score where the scores closer to 1 indicate a higher relevance.

\subsection{Unsupervised Approach}
\label{sec:method:sim-apprch}

We expect relevant text snippets to have a higher word overlap and semantic similarity with the corresponding query.
For example, for a user query:
\begin{quote}
    \textbf{Query}: {\em I am looking for a place that serves vegan food and also allows dogs inside.}
\end{quote}
the two relevant snippets have overlapping vocabulary:
\begin{quote}
\begin{enumerate}
    \item \textbf{Special diets}: vegan friendly
    \item \textbf{Review}: It was such a happy surprise that they allowed dogs inside their premises. Fanta was woofing with delight.
\end{enumerate}
\end{quote}

We map the user query ($Q_i$) and each snippet ($S_j$) for all of the restaurants into a fixed-sized vector using a mapping function $M$. We then compute the cosine similarity score between the user request and each snippet:
$$
Score(Q_i, S_{j}) = cos(M(Q_i), M(S_{j}))
$$
The cosine score estimates the distance between the query $Q_i$ and snippet $S_j$.

\vspace*{5px}
\noindent \textbf{Cos-TF-IDF: Cosine similarity with TF-IDF Encoding }\\
As a baseline, we use Term Frequency-Inverse Document Frequency(TF-IDF), an efficient and simple algorithm for matching words in a query to documents that are relevant to that query, to encode the query and the knowledge snippets~\cite{salton1988term}.
Text snippet for each restaurant is taken as a separate document to calculate TF-IDF for each word in the snippet. TF-IDF score for each word in the query is also calculated.
Then, the words in the query and the words in the text snippet are replaced by their TF-IDF score to form vocabulary-sized vectors.\footnote{The vocabulary size in our domain is 30124.}  The cosine score corresponds to the vocabulary overlap between $Q_i$ and $S_j$.

\vspace*{5px}
\noindent \textbf{Cos-BERT: Cosine similarity using BERT encoding} \\
 We use pretrained Sentence-BERT (SBERT) model to encode the query and the snippets~\cite{reimers2019sentence}. SBERT uses a pooling operation to obtain the mean of all output vectors of BERT, fine-tuned on the SNLI corpus, to derive a fixed-sized sentence embedding~\cite{devlin2019bert}.  The cosine score corresponds to the semantic distance between $Q_i$ and $S_j$.

\subsection{Supervised Approach}

We build a classifier using a DNN with a single fully-connected linear layer that takes as input encoded representation of a query and a snippet. The input 
($<CLS>query<SEP>snippet$) 
is encoded with the pretrained BERT model and the encoding of the special symbol ($CLS$) is passed into a linear layer that outputs a 2- or 3-way classification. The model is trained to minimize cross-entropy on the training set.

\vspace*{5px}
\noindent \textbf{Trans-BERT: Transfer Learning Approach}\\
\label{sec:model:class-bert}
For transfer learning we use Stanford Natural Language Inference (SNLI) and DSTC9 datasets~\cite{10.5555/3298023.3298238,kim-etal-2020-beyond}. SNLI corpus is a collection of 570,000  sets of premises and hypotheses sentences annotated with the labels {\em contradiction}, {\em entailment}, and {\em neutral} as in the following example: 
\begin{quote}
    \textbf{Premise}: A boy is jumping on skateboard in the middle of a red bridge.\\
    \textbf{Entailment}: The boy does a skateboarding trick.\\
    \textbf{Contradiction}: The boy skates down the sidewalk.\\
    \textbf{Neutral}: The boy is wearing safety equipment.\\
\end{quote}
\vspace{-5mm}
The intuition behind using a model trained on the SNLI dataset is that the relevant snippet for the user's query would be classified as {\em entailment} while the irrelevant snippets would be classified as {\em neutral} or {\em contradiction}. Hence, we train a 3-way classification and use the score of {\em entailment} class to estimate relevance.

DSTC9 dataset was constructed for the purpose of extracting answers to follow-up questions in a dialogue system. It contains 938 question and answer pairs for the train, hotel, and restaurant domains as shown in following examples.
\begin{quote}
    \textbf{Request}: Are children welcomed at this location?\\
    \textbf{Reply}: Yes, you can stay with children at A and B Guest House.\\\\
    \textbf{Request}: Can my small dog stay with me?\\
    \textbf{Reply}: Pets are not allowed at the A and B Guest House.\\
\end{quote}
\vspace{-5mm}
We use the question-answer pairs as the positive examples (relevant) for training the model.  The negative examples for training a binary classifier are extracted by randomly sampling answers from different questions following the approaches used by the authors.

\vspace*{5px}
\noindent \textbf{Dom-BERT: In-domain Training Approach }\\
The models that are trained on in-domain data usually achieve better performance. To compare the models trained with the in-domain data with the transfer/unsupervised approaches, we train a supervised binary classifier on ResQS and the combination of ResQS and DSTC9 datasets. A pre-trained BERT encoder was fine-tuned on the in-domain classification task using the pruning method  that removes the task-irrelevant neural connections from the BERT model to better reflect the data for the task~\cite{DBLP:journals/corr/abs-2105-03343}.

\section{Experiments and Results}
\label{sec:exp}
Evaluating the overall relevance of a recommended restaurant is challenging, as the information presented to the user would bias their judgement. We assume that the restaurants preferred by the  user are associated with the snippets that are relevant to the query. The models are evaluated based on the scores they assign to the snippets. 

 We  use the two unsupervised and the transfer learning from SNLI models for {\em information retrieval} to extract a recommended restaurant for each query by ranking restaurants based on the relevance score of the snippets (see~\cref{sec:exp:info-ret}). The dataset produced with the information retrieval is then manually annotated and used to train supervised  {\em snippet relevance classification} models.
To compare snippet relevance classification methods, we use a threshold to classify each snippet's relevance to the query and compute the classifier's precision, recall, and F1 (see \cref{sec:exp:rel-score}).

\subsection{Information Retrieval}
\label{sec:exp:info-ret}

\begin{table}[!t]
\centering
\caption{Snippet Relevance  and the overall recommendation  of information retrieval.}
\resizebox{\textwidth}{!}{
\begin{tabular}{|c|c|c|c|c|c|c|}
\hline
\multirow{2}{*}{\textbf{Model}} & \multicolumn{4}{c|}{\textbf{Snippet Relevance }}                                                                                                                                                                                          & \multicolumn{2}{c|}{\textbf{Overall Recommendation}}                                                                                                                                  \\ \cline{2-7} 
                       & \begin{tabular}[c]{@{}c@{}}~~~~All~~~~\\~~~~(500)~~~~~\end{tabular} & \begin{tabular}[c]{@{}c@{}}Menu Item\\ (230)\end{tabular} & \begin{tabular}[c]{@{}c@{}}Objective\\ (125)\end{tabular} & \begin{tabular}[c]{@{}c@{}}Subjective\\ (155)\end{tabular} & \begin{tabular}[c]{@{}c@{}}\% with at least one\\ relevant snippet\end{tabular} & \multicolumn{1}{l|}{\begin{tabular}[c]{@{}l@{}}avg number of \\ relevant snippets\end{tabular}} \\ \hline
Cos-TF-IDF                 & \textbf{65 \%}                                      & 59\%                                                      & 53\%                                                      & \textbf{67\%}                                              & 83\%                                                                     & \textbf{3.25}                                                                                     \\ \hline
Cos-BERT               & 57 \%                                               & \textbf{61\%}                                             & \textbf{55\%}                                             & 60\%                                                       & \textbf{86\%}                                                            & 2.86                                                                                              \\ \hline
Transfer-BERT (SNLI)   & 16 \%                                               & 10\%                                                      & 13\%                                                      & 20\%                                                       & 41\%                                                                     & 0.79                                                                                              \\ \hline
\end{tabular}
}
\label{tab:exp:snip-relevance-result}
\end{table}

For each  of the queries we apply unsupervised and transfer learning methods to score each snippet and use these scores to rank the restaurants. As a dialogue system can present multiple options in response to a query, we randomly select one of the top five recommended restaurants and manually annotate the top five matching snippets  with a binary label {\it relevant/not-relevant} (see~\cref{sec:data:annotation}).  
 For the two unsupervised cosine-similarity (Cos-TF-IDF and Cos-BERT) and transfer (Trans-BERT-SNLI) methods we report the {\it snippet relevance} (the percent of snippets extracted by the model  labelled as {\it relevant}) and the {\it overall recommendation} quality
  (see~\cref{tab:exp:snip-relevance-result}). 
We further analyze the relevance scoring performance across three query types that mention a menu item, objective, or subjective information.

Cosine similarity approach using TF-IDF encoding (Cos-TF-IDF), which relies on exact keyword match, achieves the highest overall snippet relevance of 65\%, followed by Cos-BERT with 57\%. Transfer learning did not work well, yielding only 16\% snippet relevance. We observe that Cos-TF-IDF performance is the highest (67\%) on the queries with {\em subjective} information that contain adjectives (`excellent', `great'). However, its performance is below Cos-BERT model on the queries with {\em objective} and {\em menu item} information (`fireplace', `dogs', 'desserts'). For a user query {\em `Find me a restaurant with great desserts'},
Cos-TF-IDF model extracts a restaurant with generic positive reviews which are not relevant to the query:
\begin{quote}
\begin{enumerate}
    \item \textbf{Review}: Great service, great food, had a great night!  Good value for money and great atmosphere, definitely coming back.
    \item \textbf{Review}: Great find in cambridge
\end{enumerate}
\end{quote}

However, Cos-BERT model extracts a restaurant with the snippets relevant to this query that focus on the quality of deserts:
\begin{quote}
\begin{enumerate}
    \item \textbf{Review}: The food in this restaurant is very good.  However, it is the desserts that steal the show.  I have sometimes been there just for dessert.
    \item \textbf{Review}: Fabulous french style food and cocktails. we plan to return for dessert only as the selection looked amazing and we were quite full from our meal.
\end{enumerate}
\end{quote}

Using BERT for embedding the query and the snippets appears to capture the semantics which is especially important for the queries with objective information. 

 The overall recommendation accuracy is the proportion of recommended restaurants that would satisfy the user. We approximate the subjective user satisfaction with 
 1) the percentage of the queries for which at least one of the top five snippets was labelled as {\em relevant} and 2) the average number of the top five snippets labeled as {\em relevant}.  
Cos-BERT model outperforms the Cos-TF-IDF on the first metric (86\% vs 83\%) and  Cos-TF-IDF outperforms Cos-BERT on the second one (3.25 vs 2.86).

Our dataset contains 62.3K snippets. Exact word match (TF-IDF) may have outperformed the semantic  method (Cos-BERT) because it was likely to find exact word match for each query. However, for smaller datasets, where vocabulary in a query may not match any of the text snippet, semantic methods may be more beneficial. 

The overall relevance scoring accuracy is 48\%, resulting in a balanced dataset of 1710 query-snippet (ResQS) pairs which we use to train and evaluate supervised relevance labeling models described in the next section.

\subsection{Snippet Relevance Classification}
\label{sec:exp:rel-score}
\begin{table}[t!]
\centering
\caption{Relevance classification performance on ResQS dataset.}
\begin{tabular}{|l|c|c|c|c|}
\hline
\textbf{Model Type}  & \textbf{Training data }                      & \multicolumn{1}{l|}{\textbf{Avg Precision}} & \multicolumn{1}{l|}{\textbf{Avg Recall}} & \multicolumn{1}{l|}{\textbf{Weighted F1}} \\ \hline \hline
\begin{tabular}[c]{@{}l@{}}Always-relevant\\  baseline\end{tabular}& - & 0.240                                   & 0.490                                &  0.322                                     \\ \hline
\multicolumn{5}{c}{Unsupervised methods} \\\hline
Cos-TF-IDF   & -  & 0.752                                   & 0.519                                & 0.365                                     \\ \hline

Cos-BERT    & - & 0.703                                   & 0.672                                & 0.661                                     \\ \hline
\multicolumn{5}{c}{Transfer learning methods} \\\hline
Trans-BERT& SNLI & 0.429                                   & 0.493                              & 0.368                                     \\ \hline
Trans-BERT& DSTC9 & 0.771                                   & 0.768                                & 0.769                                     \\ \hline
\multicolumn{5}{c}{Supervised in-domain training methods} \\\hline
Dom-BERT & ResQS & 0.829                                   & 0.825                                & 0.824                                     \\ \hline
Dom-BERT &  ~~~ResQS + DSTC9~~~ & \textbf{0.859}                          & \textbf{0.857}                       & \textbf{0.856}                            \\ \hline
\end{tabular}
\label{tab:sec:experimets:supervised}
\end{table}

We compare the snippet relevance classification performance of the unsupervised (Cos-TF-IDF, Cos-BERT), transfer (Trans-BERT) and supervised (Dom-BERT) methods. The unsupervised cosine similarity methods use a threshold of 0.5 to determine relevance of a snippet to a query. 
The Trans-BERT models are trained on the publicly available SNLI and DSTC9 datasets. 
The model trained on SNLI dataset outputs a 3-way classification with the classes {\em contradiction}, {\em entailment}, and {\em neutral}. To apply it on our binary query relevance detection task, we combine {\em contradiction} and {\em neutral} into the {\em not relevant} class and use {\em entailment} as the {\em relevant} class. 
The unsupervised and transfer methods  are evaluated on the full ResQS corpus as they do not use any of its data for training and the supervised methods are evaluated with 10-fold cross-validation.

\cref{tab:sec:experimets:supervised} shows the average precision, recall, and weighed F1 scores on the ResQS dataset. Cos-TF-IDF method achieves F1 of 0.365, slightly higher than the baseline that predicts all snippets as {\em relevant} (F1=0.322) while Cos-BERT  achieves F1 of 0.661. We observe that Cos-TF-IDF which relies on word match has a lower recall than Cos-BERT which captures semantic meaning (0.519 vs 0.672). 

The Transfer-BERT trained on SNLI achieves F1 of 0.368, slightly higher than the {\it always-relevant} baseline but lower than the unsupervised Cos-BERT model. The Trans-BERT model trained on DSTC9, a dataset more closely resembling to the target data,  achieves F1 of 0.769  outperforming both of the unsupervised methods.
The model trained on in-domain ResQS dataset achieves F1 of 0.824 and outperforms all unsupervised and transfer learning models. The best result (F1=0.856) is obtained by training on the combined in-domain ResQS and DSTC9 dataset. 

\subsection{Discussion}

Our aim was to collect data that covers most of the aspects of the individual restaurants. In addition to the standard aspects (cuisine, location) provided by the restaurants, we also used reviews which augmented the restaurant information with further non-standard (fireplace, dogs, exotic) aspects thus providing a comprehensive unstructured dataset for handling unconstrained queries.

We then test how accurately an unsupervised method can extract relevant items using  unstructured data. Cosine similarity with the BERT encoder  method resulted in 57\% relevant snippets on the information retrieval task and achieved F1 score of .661 on the snippet relevance classification task. With this approach 86\% of top-5 recommended restaurants had at least one relevant snippet (the estimated recommendation accuracy). Assuming that an item with more relevant snippets is a better match for a user query, we expect to achieve even higher restaurant recommendation accuracy using a more accurate supervised snippet relevance scoring model.  The supervised model achieves F1 score of 0.856 on the snippet relevance classification task.

Although we achieve improvements with the supervised models, it should be noted that they are slower compared to unsupervised methods\footnote{Experiments were conducted with i7-6 cores CPU and single GTX 1080 GPU.}. If using a supervised model to extract relevant information from unstructured data is not real-time, it becomes unfeasible for use in a dialogue system. To reduce latency while maintaining accuracy, we could use a combination of models by first filtering snippets with an unsupervised method and then applying a supervised model on a smaller dataset.

\section{Conclusions}
\label{sec:conclusions}
In this work, we aim to improve naturalness of interaction with an information navigation dialogue system. 
When a user is not primed by instructions, most user queries in the restaurant search domain include out-of-schema search preferences.  Such queries cannot be handled by a purely schema-driven dialogue system, resulting in ineffective and unnatural conversations. 
To address this problem, we propose an entity retrieval method by incorporating a snippet relevance classifier into the pipeline of a schema-driven dialogue system.

{\bf We present an effective methodology for extending schema-driven dialogue systems to use unstructured knowledge and handle out-of-schema user preferences.}\footnote{The project was completed during during a four month internship and the annotation costs were under \$300.}   
We update the Cambridge restaurants database, and extend it with text knowledge snippets. 
To simulate a naive user, we collect restaurant queries from users without biasing them with specific instructions
In our experimental restaurant search domain, the snippets were obtained from publicly available restaurant reviews and descriptions and annotated by Amazon Mechanical Turk workers. We build a supervised text relevance classification model on the annotated data and compare its performance with an unsupervised method. 
We show that on the annotated dataset an {\bf unsupervised/supervised classifier  achieves a weighted F1 of .661/.856}.
In future work, we will incorporate the proposed approach into the Cambridge restaurant search dialogue system and evaluate it with users.

\bibliography{refs}
\bibliographystyle{spmpsci}

\end{document}